\documentclass{article} 
\usepackage[submission]{colm2025_conference}
\colmpreprinttrue

\usepackage{microtype}
\usepackage{hyperref}
\usepackage{url}
\usepackage{booktabs}
\usepackage{algorithm}
\usepackage{algorithmic}
\usepackage{lineno}
\usepackage{graphicx}
\usepackage{xspace}
\usepackage{caption}
\usepackage{multirow} 
\usepackage{authblk}

\definecolor{darkblue}{rgb}{0, 0, 0.5}
\hypersetup{colorlinks=true, citecolor=darkblue, linkcolor=darkblue, urlcolor=darkblue}

\usepackage{amsmath,amsfonts,bm}

\def\eqref#1{equation~\ref{#1}}

\def\1{\bm{1}}

\def\rmA{{\mathbf{A}}}
\def\rmB{{\mathbf{B}}}
\def\rmC{{\mathbf{C}}}

\def\rmK{{\mathbf{K}}}

\def\rmO{{\mathbf{O}}}

\def\rmQ{{\mathbf{Q}}}

\def\rmV{{\mathbf{V}}}
\def\rmW{{\mathbf{W}}}
\def\rmX{{\mathbf{X}}}

\def\vo{{\bm{o}}}

\def\vx{{\bm{x}}}
\def\vy{{\bm{y}}}

\DeclareMathAlphabet{\mathsfit}{\encodingdefault}{\sfdefault}{m}{sl}
\SetMathAlphabet{\mathsfit}{bold}{\encodingdefault}{\sfdefault}{bx}{n}

\newtoggle{showcomments}
\toggletrue{showcomments}

\newcommand{\ssmname}{\textbf{M1}}
\newcommand{\baseline}{DeepSeek-R1-Distill-Qwen-1.5B}

\title{M1: Towards Scalable Test-Time Compute with \\ Mamba Reasoning Models}

\author{%
  Junxiong Wang\textsuperscript{1}, 
  Wen-Ding Li\textsuperscript{2}, 
  Daniele Paliotta\textsuperscript{3}\thanks{Work done when  interned at TogetherAI}\\[1ex]
  Daniel Ritter\textsuperscript{2}, 
  Alexander M. Rush\textsuperscript{2}, 
  Tri Dao\textsuperscript{1,4}
}

\affil{\textsuperscript{1}TogetherAI, \textsuperscript{2}Cornell University, \textsuperscript{3}University of Geneva, \textsuperscript{4}Princeton University}

\begin{document}


\maketitle

\begin{abstract}
Effective reasoning is crucial to solving complex mathematical problems. Recent large language models (LLMs) have boosted performance by scaling test-time computation through long chain-of-thought reasoning. However, transformer-based models are inherently limited in extending context length due to their quadratic computational complexity and linear memory requirements. 
In this paper, we introduce a novel hybrid linear RNN reasoning model, \ssmname{}, built on the Mamba architecture, which allows memory-efficient inference. Our approach leverages a distillation process from existing reasoning models and is further enhanced through RL training. 
Experimental results on the AIME and MATH benchmarks show that \ssmname{} not only outperforms previous linear RNN models but also matches the performance of state-of-the-art Deepseek R1 distilled reasoning models at a similar scale.
We also compare our generation speed with a highly performant general purpose inference engine, vLLM, and observe more than a 3x speedup compared to a same size transformer. With throughput speedup, we are able to achieve higher accuracy compared to DeepSeek R1 distilled transformer reasoning models under a fixed generation time budget using self-consistency voting. Overall, we introduce a hybrid Mamba reasoning model and provide a more effective approach to scaling test-time generation using self-consistency or long chain of thought reasoning. Code and pre-trained checkpoints are open-sourced at \href{https://github.com/jxiw/M1}{github.com/jxiw/M1}.
\end{abstract}

\section{Introduction}

Robust and effective reasoning is the cornerstone for successfully performing tasks in domains such as mathematics and programming. Additionally, performance on reasoning tasks can often be boosted by generating longer sequences and/or generating many sequences in parallel~\citep{snell2024scalingllmtesttimecompute}. However, current transformer-based large language models (LLMs) face significant challenges when tasked with processing long sequences with large batch sizes. These models are constrained by a quadratic increase in computational complexity as the sequence length grows, coupled with a linear escalation in memory requirements. This combination makes it increasingly difficult for models to scale efficiently when handling large inputs.


Although linear hybrid RNN models~\citep{gu2024mambalineartimesequencemodeling,dao2024transformersssmsgeneralizedmodels,beck2024xlstmextendedlongshortterm,yang2024gatedlinearattentiontransformers,peng2023rwkvreinventingrnnstransformer} have shown great potential as an alternative to transformer-based on general language models, their effectiveness on reasoning tasks remains unclear. Since modern reasoning models typically generate long chains of thought for challenging math questions, it is uncertain whether the performance of hybrid linear RNNs diminishes in such scenarios.

In this paper, we propose \ssmname{} and show that it is possible to derive strong hybrid reasoning models by efficiently transferring reasoning capabilities from a large transformer model. Our training process involves distilling knowledge, incorporating math and reasoning abilities through supervised fine-tuning (SFT), and finally, boosting performance using reinforcement learning (RL) training. In total, the training process requires fewer than 50 billion tokens. In contrast, \baseline{} is finetuned from Qwen2.5 MATH 1.5B which is trained using over 1 trillion MATH tokens on top of Qwen2.5. 




We demonstrate that our hybrid models achieve a 3x speedup compared to transformers of the same size when served using a highly performant general purpose inference engine, vLLM, at large batch sizes. This gain is mainly due to large batches and long sequences, decoding being generally memory-bound. Lower memory usage of hybrid models can transform this advantage into a speed gain. The decoding speedup is approximately linear with the volume of model's memory access \citep{yuan2025native}.

Notably, this speedup can be converted to a gain in reasoning accuracy. Studies~\citep{snell2024scalingllmtesttimecompute,li2025surveyllmtesttimecompute,chen2025towards} show that techniques such as self-consistency~\citep{wang2023selfconsistencyimproveschainthought} and verification~\citep{cobbe2021trainingverifierssolvemath} at test time can significantly boost model reasoning performance. Under these conditions, a high-throughput model can further enhance its performance by generating more samples.




The paper is organized as follows. Section~\ref{sec:related} covers related work,  Section~\ref{sec:hybrid_reason} introduces our pipeline for distilling a hybrid reasoning model, and Section~\ref{sec:experiments} presents our results evaluating \ssmname{} on math benchmarks. Sections \ref{sec:speed_eval} and \ref{sec:test_time_scaling} evaluate the performance gains of \ssmname{} in terms of both inference speed and scaling test-time compute. Section~\ref{sec:analysis} provides some additional analysis of the impact of different generation lengths when training on RL, and of the impact of the different steps of the distillation pipeline we propose on performance. 

Overall, we show that \ssmname{} performs on par with \baseline{}, achieving scores of 82 on MATH500~\citep{hendrycks2021measuringmathematicalproblemsolving}, 23 on AIME25~\citep{aime25}, 28 on AIME24~\citep{aime24}, and 47 on OlympiadBench~\citep{he2024olympiadbench}, while offering 3x faster inference throughput, even compared to the highly optimized vLLM~\citep{kwon2023efficient} implementation for Transformer models.

\section{Related Work} 
\label{sec:related}

\subsection{Reasoning models}
Recent models like Deepseek-R1 ~\citep{deepseekai2025deepseekr1incentivizingreasoningcapability} have shown the potential of RL training to improve performance on verifiable reasoning tasks, such as math problem solving and programming. Additional work has proposed methods for inducing this reasoning behavior via supervised fine-tuning, either on curated data ~\citep{muennighoff2025s1simpletesttimescaling} or on generated pairs of traces ~\citep{yang2025thinkingpreferenceoptimization}. Other approaches also combine search procedures such as MCTS with language models ~\citep{qi2024mutualreasoningmakessmaller} or alter standard RL training schemes to control the length of generated outputs ~\citep{aggarwal2025l1controllinglongreasoning}. After training, these models solve complex tasks by generating long chains of thought, which often include subtasks of the overall problem, multiple attempted solutions, and backtracking over prior attempts~\citep{gandhi2025cognitivebehaviorsenableselfimproving}. Since the performance of these models, both during training and inference, relies on generating lengthy chains of thought, more efficient architectures can enable larger scale training and less costly generation. 

\subsection{Enhancing Reasoning via Scaled Inference Compute} Increasing the computational budget during inference has become a promising approach to boost LLM performance. Methods like Chain of Thought (CoT) and its derivatives have achieved notable gains on reasoning benchmarks by breaking down complex tasks into intermediate steps~\citep{wei2023chainofthoughtpromptingelicitsreasoning, yao2023treethoughtsdeliberateproblem}. Although decomposing tasks improves reasoning, it also lengthens generation sequences and raises computational costs. Some recent studies even indicate that this extra computation might itself enhance model capabilities~\citep{pfau2024letsthinkdotdot}. In addition, adaptive compute allocation during inference has been explored. For example, \citet{goyal2024thinkspeaktraininglanguage} incorporated pause tokens into the vocabulary, allowing models to distribute compute more efficiently and improve both reasoning and overall task performance. LightTransfer~\citep{zhang2024lighttransfer} introduces a lightweight method that detects lazy layers and replaces their full attention with streaming attention—slashing KV-cache overhead and boosting throughput.

Another strategy involves generating several outputs and selecting the best one. Researchers have developed various sampling algorithms to diversify and enhance the quality of generated responses, thereby increasing the chances of retrieving the most accurate answer~\citep{wang2023selfconsistencyimproveschainthought, renze2024effectsamplingtemperatureproblem, zhang2023planninglargelanguagemodels}. Moreover, outcome and process reward models (ORMs and PRMs) have been introduced to evaluate responses and steer intermediate generation steps~\citep{lightman2023letsverifystepstep, zhang2024restmctsllmselftrainingprocess, luo2024improvemathematicalreasoninglanguage, uesato2022solvingmathwordproblems}.

Recent investigations reveal that, under fixed compute budgets, smaller LLMs augmented with inference-time compute techniques (such as majority voting or PRM-guided search) can outperform larger models~\citep{snell2024scalingllmtesttimecompute, wu2024inferencescalinglawsempirical, beeching2024scalingtesttimecompute}. However, these results are mainly confined to Transformer-based architectures, leaving open questions about whether similar scaling laws hold for subquadratic architectures, which offer faster inference but might compromise on expressiveness.

\subsection{Alternatives to Transformer Architectures} Even though most reasoning models are based on the Transformer architecture~\citep{grattafiori2024llama3herdmodels, qwen2025qwen25technicalreport}, alternatives have been proposed to alleviate their high computational cost. Models built on top of RNNs~\citep{beck2024xlstmextendedlongshortterm, peng2023rwkvreinventingrnnstransformer}, state space models (SSMs)~\citep{gu2022efficientlymodelinglongsequences, gu2024mambalineartimesequencemodeling}, and linear attention mechanisms~\citep{katharopoulos2020transformersrnnsfastautoregressive, yang2024gatedlinearattentiontransformers} demonstrate superior inference and memory efficiency, particularly for long-context tasks and large-batch generation. The Mamba series (Mamba-1 and Mamba-2) notably introduced selective state spaces to enable linear-time sequence modeling with strong performance~\citep{gu2024mambalineartimesequencemodeling, dao2024transformersssmsgeneralizedmodels}. In addition, hybrid architectures that combine a few self-attention layers with subquadratic layers (e.g., Mamba) have emerged, showing advantages over both pure Transformer and pure subquadratic designs~\citep{lieber2024jambahybridtransformermambalanguage, ren2024sambasimplehybridstate}. Such architectures are particularly suited to meet the high compute demands of inference-time scaling, and our work investigates their scaling properties.

\subsection{Knowledge Distillation Strategies} Knowledge distillation has proven to be an effective means of transferring capabilities from large teacher models to smaller, more efficient student models~\citep{hinton2015distillingknowledgeneuralnetwork}. In LLMs, this process compresses a larger pre-trained model into a more compact version while preserving core knowledge and functionality~\citep{gu2024minillmknowledgedistillationlarge, xu2024surveyknowledgedistillationlarge}. Although larger models tend to exhibit superior reasoning abilities due to scaling properties~\citep{xu2025largereasoningmodelssurvey, wei2022emergentabilitieslargelanguage}, distillation techniques have enabled smaller models to achieve competitive reasoning performance~\citep{deepseekai2025deepseekr1incentivizingreasoningcapability, bespoke_stratos}. While most efforts have focused on intra-architecture distillation (e.g., Transformer-to-Transformer), recent studies have ventured into cross-architecture distillation. For instance, pretrained Transformers have been distilled into architectures such as RNNs~\citep{kasai2021finetuningpretrainedtransformersrnns, mercat2024linearizinglargelanguagemodels}, linear attention models~\citep{zhang2024hedgehogporcupineexpressive,zhanglolcats}, convolutional networks~\citep{ralambomihanta2024scavenginghyenadistillingtransformers}, and SSMs~\citep{bick2024transformersssmsdistillingquadratic, junxiongdaniele2024mil,paliotta2025thinking}. Whether the robust reasoning abilities of Deepseek R1~\citep{deepseekai2025deepseekr1incentivizingreasoningcapability} distilled models can be effectively transferred across different architectures remains an open question.

\section{The M1 Reasoning Model}
\label{sec:hybrid_reason}

In this section, we present a multi-stage process for building our hybrid linear RNN reasoning model, M1. The approach has three stages: distillation, SFT, and RL. We begin by distilling a Transformer model into a Mamba architecture, adapting the method of \cite{wang2025mamballamadistillingaccelerating}, which initializes the hybrid model’s weights from a transformer model. We then perform math-specific supervised fine-tuning (SFT) on general mathematical datasets to enhance the model's mathematical performance, first without yet incorporating datasets generated by reasoning-focused models, and then with reasoning data leveraging multiple large-scale datasets generated by the R1 model series. Finally, we apply R1's GRPO method to further enhance the model's math reasoning capability.


\paragraph{Stage 1: Distillation.} 

The first step in building our \ssmname{} model is distilling a pretrained transformer model into a Mamba model. We adapt the distillation approach introduced by ~\citet{wang2025mamballamadistillingaccelerating}.

\begin{algorithm}[tb]
    \small
    \label{alg:group_ssm}
    \begin{algorithmic}[1]
        \STATE \textbf{Shapes:} $B$ - Batch, $L$ - Length, $D$ - embed size, $N = D  / \text{Attention\_heads}$,$N'$ - expand  
        \STATE \textbf{Input:} $\vo_t$: (B,  D)
        \STATE \textbf{Output:} output: (B, D) 
        \STATE \textbf{New Params:} MLP, $\rmA$  
        \FOR{ each head $\rmW^K, \rmW^Q, \rmW^V, \rmW^o : (N, D)$ \\
             \hspace{1cm} after expanding to same dimension}
            \STATE \textbf{Head Parameter:} $\rmA : (N, N')$
            \STATE \text{for all positions $t$:}
            \STATE $\ \vx_t : (B, N) \gets \rmW^V \vo_t$
            \STATE $\ \rmB_t : (B, N) \gets \rmW^K \vo_t$ 
            \STATE $\ \rmC_t : (B, N) \gets \rmW^Q \vo_t$
            \STATE $\ \Delta_t : (B, N') \gets \text{MLP}(\vx_t)$
            \STATE $\overline{\rmA}_{1:T}, \overline{\rmB}_{1:T}, \overline{\rmC}_{1:T}: (B, N, N') \gets \textsc{Disc}(\rmA, \rmB, \rmC, \Delta)$
            \STATE $\vy \gets \textsc{LinearRNN}(\overline{\rmA}, \overline{\rmB}, \overline{\rmC}, \vx)$
            \STATE output $\gets \text{output} + \rmW^{O\top} \vy$
        \ENDFOR{}
        \STATE \textbf{return} output
    \end{algorithmic}
    \caption{Initializing \textsc{MambaInLlama}}
\end{algorithm}

The \textsc{MambaInLlama} framework~\citep{wang2025mamballamadistillingaccelerating} proposes distilling hybrid Transformer-Mamba models by reusing weights from attention layers. In this distillation procedure, outlined in Algorithm 1, linear projections for $\rmQ$, $\rmK$, $\rmV$, and $\rmO$ are initialized from the corresponding projections for $\rmC$, $\rmB$, $\rmX$, and $\rmO$, respectively. The newly introduced parameters in the Mamba layers are the sampling rate $\Delta$ and the dynamic parameter $\rmA$, which control the resulting Mamba module via a discretization function. Specifically, the sampling rate $\Delta \in \mathbb{R}^{N'}$ discretizes $\rmB_t, \rmC_t \in \mathbb{R}^{N \times 1}$, yielding $\overline{\rmB}_t, \overline{\rmC}_t \in \mathbb{R}^{N' \times N \times 1}$, as detailed in Algorithm 1. Different from \citet{wang2025mamballamadistillingaccelerating}, we introduce two additional linear layers to project from \texttt{head\_dim * kv\_head} to \texttt{head\_dim * n\_head}. This is because GQA ~\citep{ainslie2023gqa} is used in the transformer model to reduce the KV cache. As Mamba does not utilize a KV cache, this expansion can increase the expressiveness of $\mathbf{B}$ and $\mathbf{X}$.

We directly reuse the MLP layers; however, unlike the original approach, we replace the attention layers with Mamba layers in a single step. Subsequently, we fine-tune the entire model to expedite the training process. The distillation step involves minimizing the token-level KL divergence, aligning the entire probability distribution of the student model, $p(\cdot; \theta)$, with the teacher model, $p(\cdot; \theta_T)$, for every candidate token at position $t$. We use the reverse KL divergence, $D_{\text{KL}}(p(\cdot; \theta) \parallel p(\cdot; \theta_T))$, as our loss function rather than the forward KL divergence. We choose the reverse KL divergence due to its mode-seeking properties, which results in improved empirical performance.


We reimplement the distillation and SFT framework using the Axolotl \footnote{\url{https://github.com/axolotl-ai-cloud/axolotl}}training framework. We apply the model chat template, mask the user prompt, and compute the loss only over the tokens generated in the assistant's output. To speed up training, we use data packing to merge different sequences into a single one until we reach the maximum sequence length which is set to $8192$. We find that data packing achieves significantly better results compared to the non-packing version in distillation for the same training steps. We use the AdamW optimizer with learning rate $1\times 10^{-5}$ with cosine decay, $\beta=(0.9, 0.95)$ and a weight decay of $0.1$. 


\paragraph{Stage 2: SFT} Following the distillation procedure, we finetune the model on a large set of math problems, OpenMathInstruct-2~\citep{toshniwal2024openmathinstruct2acceleratingaimath}. As in the distillation stage, we apply the chat template to the prompts, mask the user prompt, and compute the loss only over the tokens generated in the assistant's output. We train for two epochs using the same optimizer as distillation.


After the initial fine-tuning stage, we finetune on an additional set of math problems and solutions generated by reasoning models. 
We collect a mixed reasoning dataset, including OpenR1-Math-220k \footnote{\url{https://huggingface.co/datasets/open-r1/OpenR1-Math-220k}}, OpenThoughts-114k-math\footnote{\url{https://huggingface.co/datasets/open-thoughts/OpenThoughts-114k}}, and ServiceNow-AI-R1-Distill\footnote{\url{https://huggingface.co/datasets/ServiceNow-AI/R1-Distill-SFT}}, Magpie-Reasoning-250K\footnote{\url{https://huggingface.co/datasets/Magpie-Align/Magpie-Reasoning-V2-250K-CoT-Deepseek-R1-Llama-70B}} for a total of 10B reasoning tokens. The first two datasets were generated from R1, while the last two was generated from the R1 distilled Qwen 32B model and R1 distilled Llama 70B model. We extended the training length to 24,576 because we found that it covers 99\% of the data items. We train the model for five epochs using the same optimizer as before but changing the peak learning rate to $6\times 10^{-6}$.


\paragraph{Stage 3: Reasoning RL.}
To further enhance performance, we integrate Mamba with a RL pipeline for further training.\footnote{We add it into the popular VeRL~\citep{sheng2024hybridflow} framework. In doing so, we addressed and resolved the CUDA graph incompatibility issues that previously arose during training with PyTorch's FSDP module. As a result, the updated framework now efficiently supports Mamba generation with CUDA graph enabled, making it 5x faster than with CUDA Graph disabled}  We use GRPO as the loss function. Differing from \citep{deepseek-math}, we remove the KL penalty term as empirically we find it destabilizes training. Additionally, we include an entropy bonus to encourage a more diverse policy. The resulting formula is,

\begin{equation}
L_{\text{GRPO}}(\theta) = \mathbb{E}_{\tau \sim \pi_{\theta_{\text{old}}}}\left[ \frac{\pi_{\theta}(a|s)}{\pi_{\theta_{\text{old}}}(a|s)} \hat{A}(s,a) \right] + \eta \, H(\pi_{\theta})
\end{equation}

where $\hat{A}(s,a)$ is the estimate of the advantage from multiple rollouts. We use a batch size of 128 and a PPO batch size of 64, which also determines the number of PPO iterations, \(\mu = 2\). We set the number of generations for each sequence to 8 and the maximum generation length to 32k. For optimization, we use the Adam optimizer with a learning rate of $1\times 10^{-6}$. We train for 50 steps, and pick the best checkpoint with the highest critic reward. We append the simple prompt \textit{"Let's think step by step and output the final answer within \textbackslash boxed\{\}"} to the end of each question in both training and evaluation.

\section{Experiments}



\paragraph{Model.} We adopt the Llama3.2-3B-Instruct models as distillation target models. 
For Mamba layers, we set the SSM state size to 16. Consequently, the
number of SSM groups after expansion is 3072/16 = 192 for the 3B model.
We use 6 interleaved attention layers among 28 total layers.

\paragraph{Evaluation Dataset.} Following common practice in evaluating reasoning models, we use a similar set of math benchmarks, including competition-level problems: MATH500~\citep{hendrycks2021measuringmathematicalproblemsolving}, AIME25~\citep{aime25}, AIME24~\citep{aime24}, AMC23~\citep{aime23}, and OlympiadBench~\citep{he2024olympiadbench}.

\paragraph{Evaluation Metrics.} Our model's performance is assessed using two key metrics: coverage and accuracy. In fields such as coding and formal proofs, where answers can be automatically verified, coverage translates directly to enhanced performance and is widely utilized~\citep{chen2021evaluating, brown2024largelanguagemonkeysscaling}. Coverage is often measured using the pass@k metric, with $k$ indicating the number of samples per problem~\citep{chen2021evaluating, brown2024largelanguagemonkeysscaling}. This metric estimates the likelihood that at least one correct solution exists among the $k$ samples. To minimize variance when calculating coverage, we employ the unbiased estimation formula from~\citet{chen2021evaluating}. Specifically, we generate $N \geq k$ total samples per task. The probability that a correct solution exists among a pool of $k$ generated samples can then be determined given the total number of correct solutions $C_i$ for each task.

\[
\text{pass@k} = \frac{1}{\text{\# of problems}} \sum_{i=1}^{\text{\# of problems}} \left( 1 - \frac{\binom{N - C_i}{k}}{\binom{N}{k}} \right)
\]

We implement this formula using a numerically stable approach as recommended by~\citet{chen2021evaluating}.

When using additional compute, we employ multiple aggregation strategies. The most straightforward method is majority voting, also known as self-consistency decoding~\citep{wang2023selfconsistencyimproveschainthought}, which takes the majority response among $k$ samples as the predicted answer, and uses that to compute the accuracy. 

\subsection{Reasoning Evaluation}
\label{sec:experiments}

\begin{table}[h]
\centering
\resizebox{\textwidth}{!}{%
\begin{tabular}{lccccc}
\toprule
Model 
  & AIME25 
  & AIME24 
  & MATH500 
  & AMC23 
  & OlympiadBench \\
\midrule
Qwen2.5-Math-7B-Instruct & -    & 13.3 & 79.8 & 50.6 & 40.7 \\
rStar-Math-7B~\citep{guan2025rstar}          & -    & 26.7 & 78.4 & 47.5 & 47.1 \\
Eurus-2-7B-PRIME~\citep{cui2025process}         & -    & 26.7 & 79.2 & 57.8 & 42.1 \\
Qwen2.5-7B-SimpleRL~\citep{zeng2025simplerl}     & -    & 26.7 & 82.4 & 62.5 & 43.3 \\
DeepSeek-R1-Qwen-1.5B    & 23.0 & 28.8 & \textbf{82.8} & \textbf{62.9} & 43.3 \\
\ssmname-3B              & \textbf{23.5} & \textbf{28.9} & 82.1 & 62.8 & \textbf{47.3} \\
\bottomrule
\end{tabular}
}
\caption{Evaluation results for \ssmname{}-3B, \baseline{} and other MATH models on MATH benchmarks}
\label{tab:math_results_bench_transposed}
\end{table}

\begin{table}[h]
\centering
\resizebox{\textwidth}{!}{%
\begin{tabular}{lcccccccccc}
\toprule
\multirow{2}{*}{Model} 
  & \multicolumn{2}{c}{AIME25} 
  & \multicolumn{2}{c}{AIME24} 
  & \multicolumn{2}{c}{MATH500} 
  & \multicolumn{2}{c}{AMC23} 
  & \multicolumn{2}{c}{OlympiadBench} \\
\cmidrule(lr){2-3} \cmidrule(lr){4-5} \cmidrule(lr){6-7} \cmidrule(lr){8-9} \cmidrule(lr){10-11}
 & Pass@1 & Maj@32 
 & Pass@1 & Maj@32 
 & Pass@1 & Maj@32 
 & Pass@1 & Maj@32 
 & Pass@1 & Maj@32 \\
\midrule
DeepSeek-R1-Qwen-1.5B 
  & 23.0 & \textbf{35.0} 
  & 28.8 & 49.2 
  & \textbf{82.8} & 91.0 
  & \textbf{62.9} & 54.2 
  & 43.3 & 80.3 \\
\ssmname-3B 
  & \textbf{23.5} & 34.6 
  & \textbf{29.0} & \textbf{50.5} 
  & 82.1 & \textbf{91.8} 
  & 62.8 & \textbf{55.0} 
  & \textbf{47.3} & 80.1 \\
\bottomrule
\end{tabular}
}
\caption{Maj@32 results comparing \ssmname{}-3B with \baseline{}.}
\label{tab:math_results_mine_vs_deepseek}
\end{table}

We evaluate our models using a temperature setting of 0.7 and a sequence length of 32k with evaluation tools in VeRL. We use 32k because it has become the standard for evaluating performance on reasoning models~\citep{deepseekai2025deepseekr1incentivizingreasoningcapability,deepscaler2025}. We report the pass@1 metric averaged over 64 runs; for majority voting, we repeat the metric calculation 100 times.

We report the accuracy of \ssmname{}-3B and \baseline{} in Table~\ref{tab:math_results_bench_transposed} and ~\ref{tab:math_results_mine_vs_deepseek}. We use the baseline \baseline{} since a 3B R1 reasoning model is still not available. Although \ssmname{}-3B has more parameters than \baseline{}, its speed is still comparable even with shorter contexts, so we believe this is a fair comparison.
Our model's performance is competitive with state-of-the-art open reasoning models in the same model size range and outperforms larger nonreasoning math transformer models.
Our model performs slightly worse on AIME24 compared to the \baseline{} model. Notably, \baseline{} is built on top of the Qwen2.5 MATH models, which were finetuned with over 1T MATH tokens on top of the Qwen2.5 models, significantly more training data than what \ssmname{}-3B used in total.























\subsection{Speed Evaluation}
\label{sec:speed_eval}

We benchmark inference time with our model against a transformer model (Llama-3.2.-3B~\citep{grattafiori2024llama3herdmodels}) of the same size. We use vLLM (version 0.6.3), which is the version used in VeRL for efficient rollouts. We also compare against DeepSeek-R1-Distill-Qwen-1.5B~\citep{deepseekai2025deepseekr1incentivizingreasoningcapability}, a reasoning transformer model that is half the size of \ssmname{}. This model has the same number of layers as the 3B parameter transformer, but the hidden dimension is half the size. 



According to~\citet{deepscaler2025}, the average generation length of reasoning models on MATH questions is 4k to 5k. We  therefore fix a decoding length of 4096 (and prompt length of 256) and benchmark our model across a range of batch sizes. We vary the batch size from 8 to 512, measuring the inference latency across different models.  

\begin{figure}[ht]
  \noindent
  \begin{minipage}[t]{0.48\textwidth}
    \centering
    \includegraphics[width=\linewidth, trim=0 0 0 0, clip]{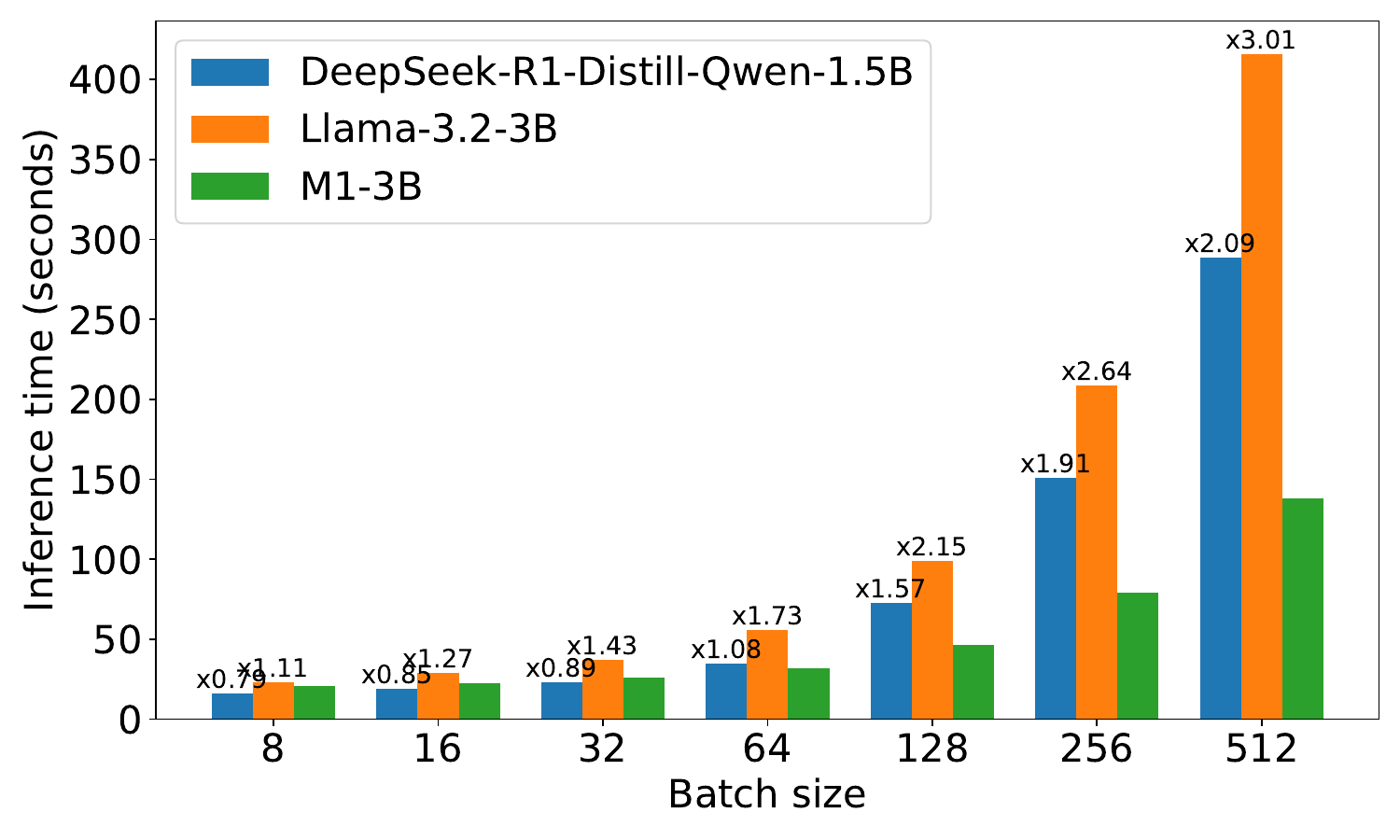}
    \captionsetup{justification=centering, skip=1pt, margin=0pt}
    \captionof{figure}{Inference latency when using prompt length 256 and decoding length 4096.}
    \label{fig:speed_gen4096}
  \end{minipage}
  \hfill
  \begin{minipage}[t]{0.48\textwidth}
    \centering
    \includegraphics[width=\linewidth, trim=0 0 0 0, clip]{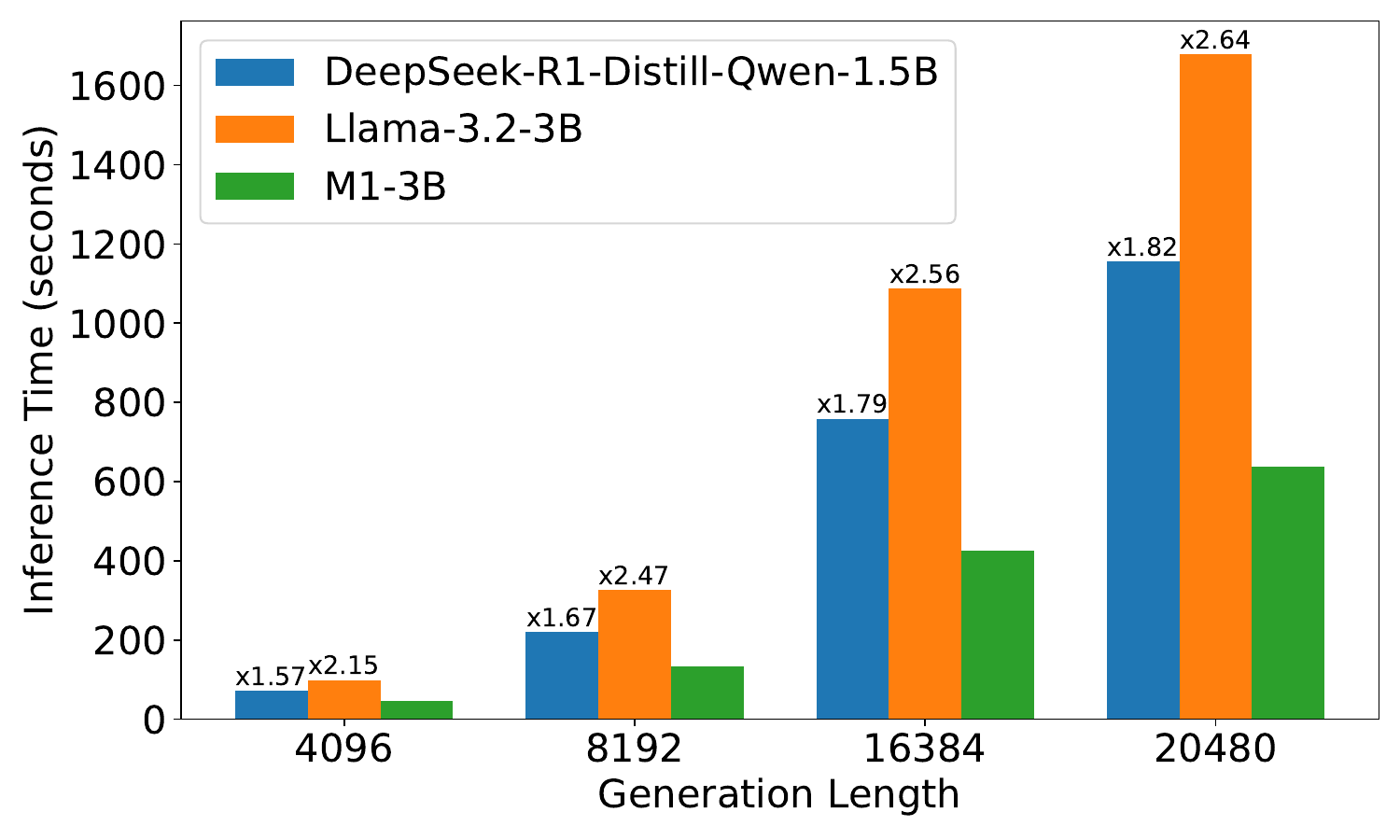}
    \captionsetup{justification=centering, skip=1pt, margin=0pt}
    \captionof{figure}{Inference latency when using batch size 128.}
    \label{fig:speed_bsz128}
  \end{minipage}
\end{figure}

We perform our benchmarking on a single NVIDIA H100 GPU with greedy decoding. To ensure that every model generates up to the set maximum number of tokens, we use \texttt{ignore\_eos=True}. Before recording results, we warm up the system with two runs. The final performance metrics are then averaged over three subsequent runs. The inference speeds of the models across batch sizes are shown in Figure \ref{fig:speed_gen4096}. \ssmname{} achieves a 3× speedup over similarly-sized transformers when using a batch size of 512 and a decoding length of 4096, demonstrating its effectiveness in large-batch generation settings.

The maximum length of generated sequences is also an important factor in RL training, as longer sequences allow the model to use more compute during learning by generating longer chains-of-thought, shown in Figure~\ref{fig:rl-training-length}. To benchmark our model in this setting, we fix the batch size to 128, and vary the generation length. We compare against the same two models as in the batch size varying case, and the results are shown in Figure \ref{fig:speed_bsz128}. As the generated sequence length increases, \ssmname{} achieves increasing speedups relative to the baseline models, and consistently generates at least 2x faster than Llama-3.2-3B (2.64x faster for the longest sequence length).

It is well-known that LLM inference comprises a prefilling (compute-bound) and a decoding (memory-bound) stage. For math reasoning models, it is common to assume that decoding takes much longer than prefilling, since prefilling only uses a short MATH question, while decoding generates long answers. Under these settings, the process is memory-bound. Given that Mamba is highly memory-efficient and we only use a SSM state size of 16, these memory advantages translate into improved speed.

\subsection{Test-Time Scaling}
\label{sec:test_time_scaling}




Given a fixed time budget, \ssmname{} can generate more sequences or longer sequences compared to a transformer model, which can hopefully boost its performance. We evaluate the effect of test-time compute scaling on model performance. We scale both the number of samples generated as well as the length of generated samples, to see if \ssmname{} benefits from additional compute along these axes. We aim to investigate whether the speed benefit from section \ref{sec:speed_eval} can translate into an accuracy gain. 

\textbf{Scaling with majority vote.}


\begin{figure}[h]
  \centering
  \begin{minipage}{0.5\textwidth}
    \centering
    \includegraphics[width=\textwidth,clip]{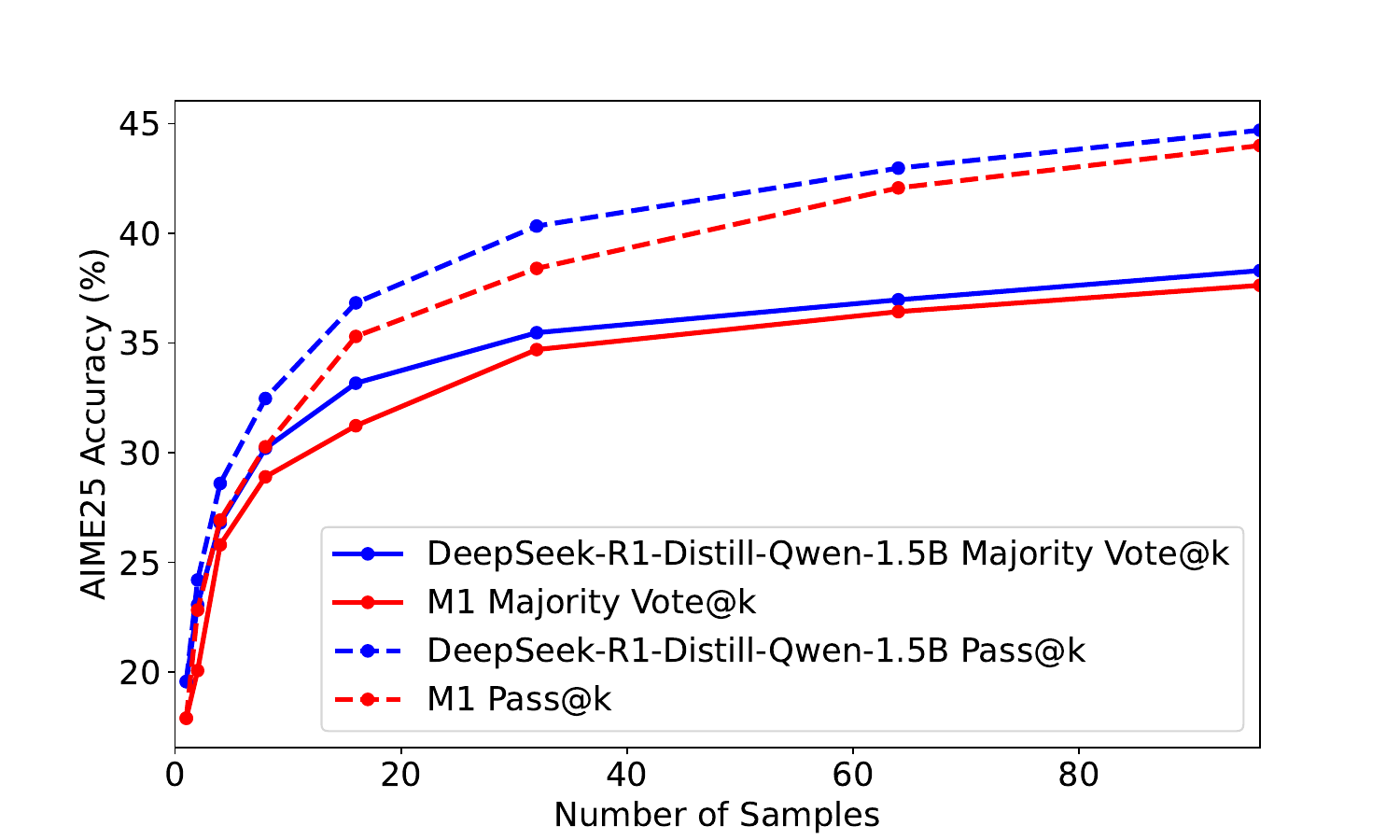}
  \end{minipage}%
  \hspace{-0.05\textwidth}%
  \begin{minipage}{0.5\textwidth}
    \centering
    \includegraphics[width=\textwidth,clip]{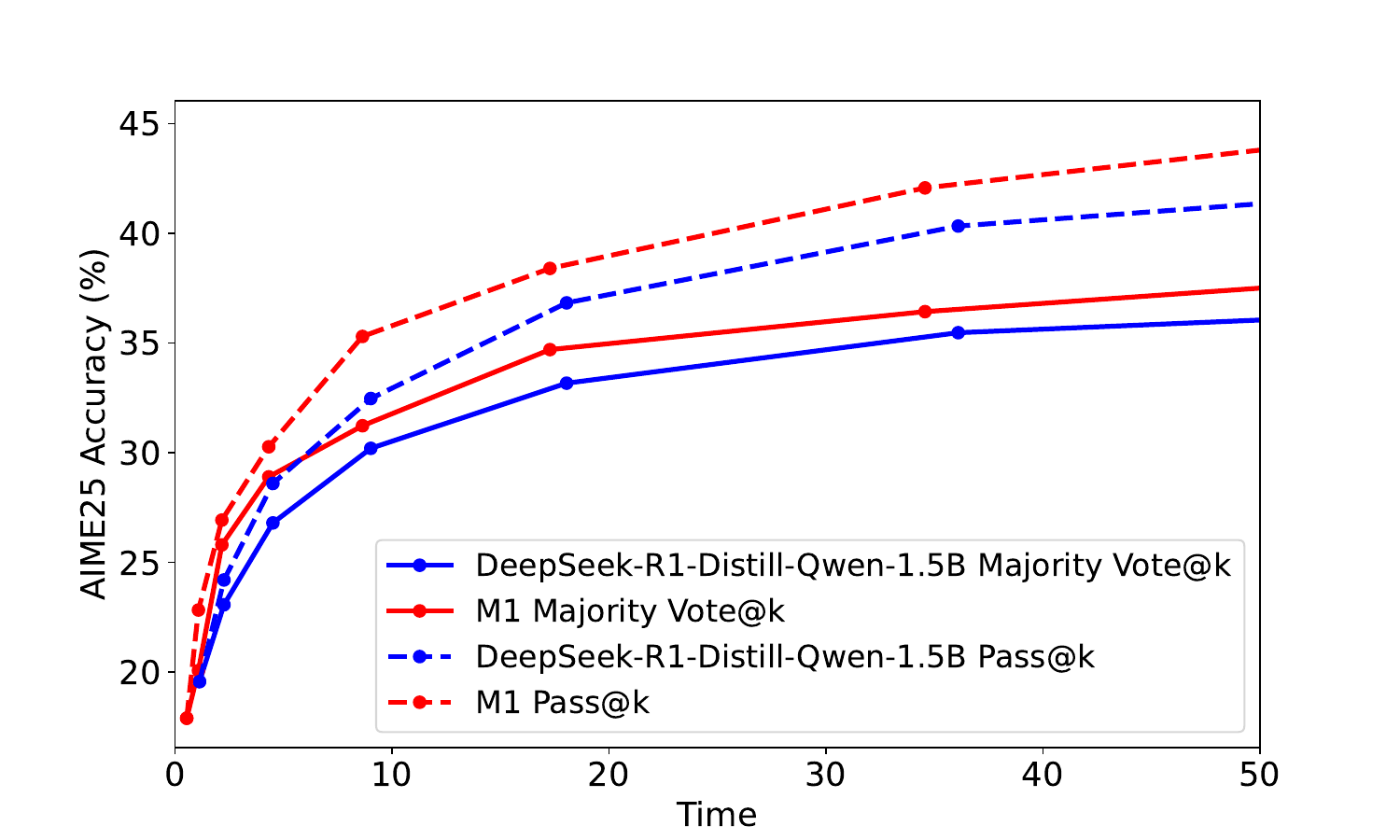}
  \end{minipage}
  \caption{Number of samples vs. AIME25 accuracy (left) and generation time (seconds) vs. AIME25 accuracy (right). Both graphs include pass@1 and majority voting accuracies for \ssmname{} and DeepSeek-R1-Distill-Qwen-1.5B.}
  \label{fig:scale_sample}
\end{figure}

The left side of Figure \ref{fig:scale_sample} shows the effect of scaling the number of generated samples (while fixing the maximum decoding length) on AIME25 accuracy. Both the baseline model and \ssmname{} see increasing accuracy as the number of samples increases, with \ssmname{} nearly matching the baseline performance for larger sample sizes. The efficient generation of \ssmname{} also means that generating large number of samples at test-time is faster than for the baseline transformer model. 

We quantify this efficiency in the right side of Figure \ref{fig:scale_sample}, which compares the number of seconds spent generating samples against the resulting accuracy. To compute the time values on the x-axis, we find an optimal throughput value (in tokens per second) for each model by increasing batch sizes until throughput decreases. The optimal values were 7263 T/s for DeepSeek-R1-Distill-Qwen-1.5B, and 15169 T/s for \ssmname{}. We then assume that each generated sample is maximum length (8K), and compute the seconds required for one sample from one model as 8K divided by the throughput. We then convert the left graph of Figure \ref{fig:scale_sample} into the right graph, by multiplying the number of samples for each datapoint by the seconds required per sample for each model. As an example, \ssmname{} requires roughly a half second (8K/15K) per sample, so the accuracy value for \ssmname{} at 32 samples on the left graph appears at approximately 16 seconds on the right graph.

\textbf{Scaling with longer sequences}

\begin{figure}[h]
  \centering
  \begin{minipage}{0.48\textwidth}
    \centering
    \includegraphics[width=\textwidth,clip]{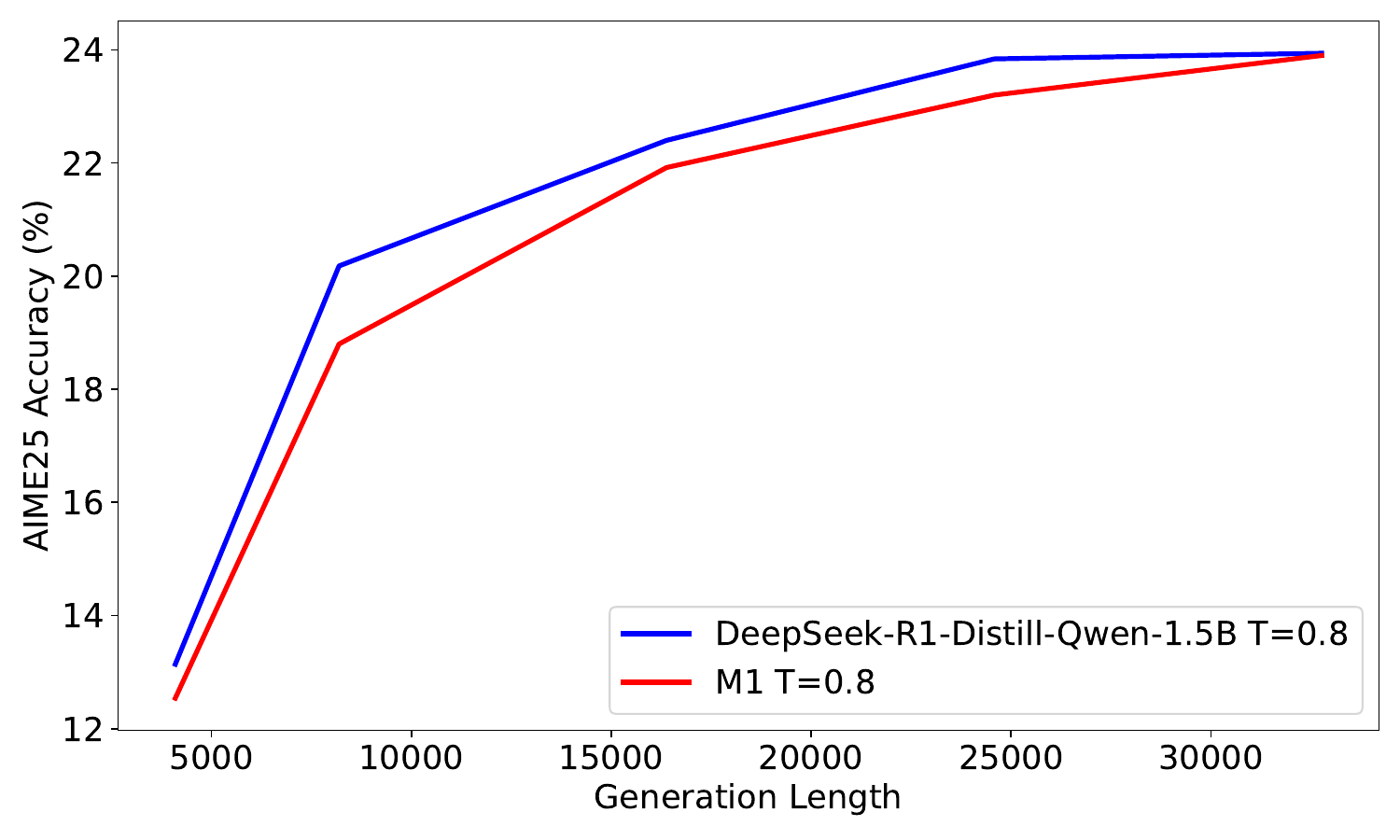}
  \end{minipage}
  \hspace{0.01\textwidth}
  \begin{minipage}{0.48\textwidth}
    \centering
    \includegraphics[width=\textwidth]{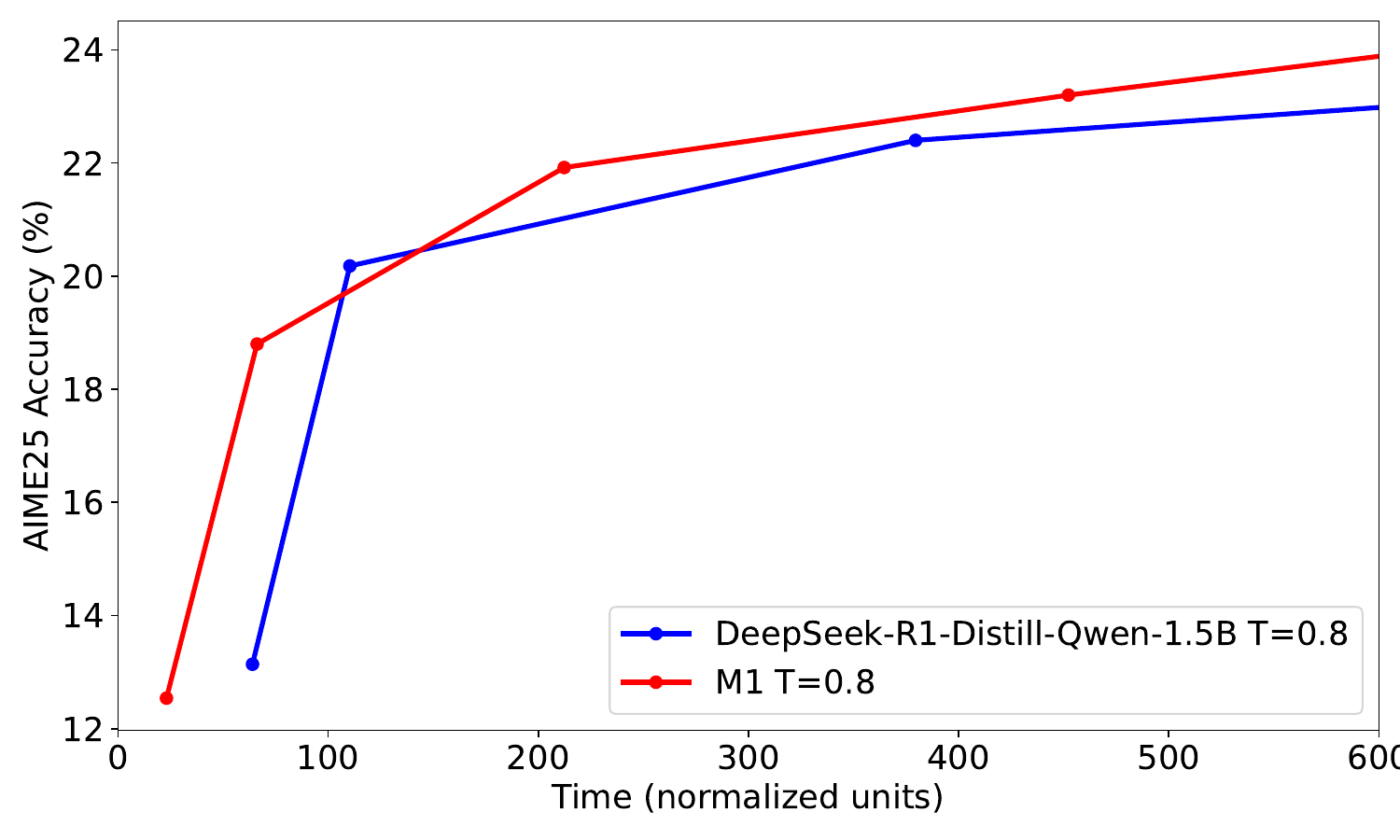}
  \end{minipage}
  \caption{Generation length vs. AIME25 accuracy (left) and generation time (seconds) vs. AIME25 accuracy (right). Sampling for both models is done using a temperature of 0.8.}
  \label{fig:scale_length}
\end{figure}

Figure \ref{fig:scale_length} shows the effect of scaling the maximum length of the generated answer, while fixing the number of generated samples to one. For both the baseline and \ssmname{}, increasing the maximum sequence length leads to increased accuracy, as shown in the left graph in Figure \ref{fig:scale_length}. After converting from generation length to the seconds required to generate (done in the same way as Figure \ref{fig:scale_sample}, but dividing the generation length by throughput), we can see the accuracy gain per time spent generating on the right side of Figure \ref{fig:scale_length}. In this case, \ssmname{} actually gets a higher accuracy for the same amount of time spent generating at 4 of the 5 evaluated sequence lengths, showing the benefits of efficient generation for test-time compute scaling.

\section{Analysis}
\label{sec:analysis}
 

\textbf{Increasing Training Length in RL boosts model performance}

\begin{figure}[h]
  \centering
  \includegraphics[width=0.45\textwidth,clip]{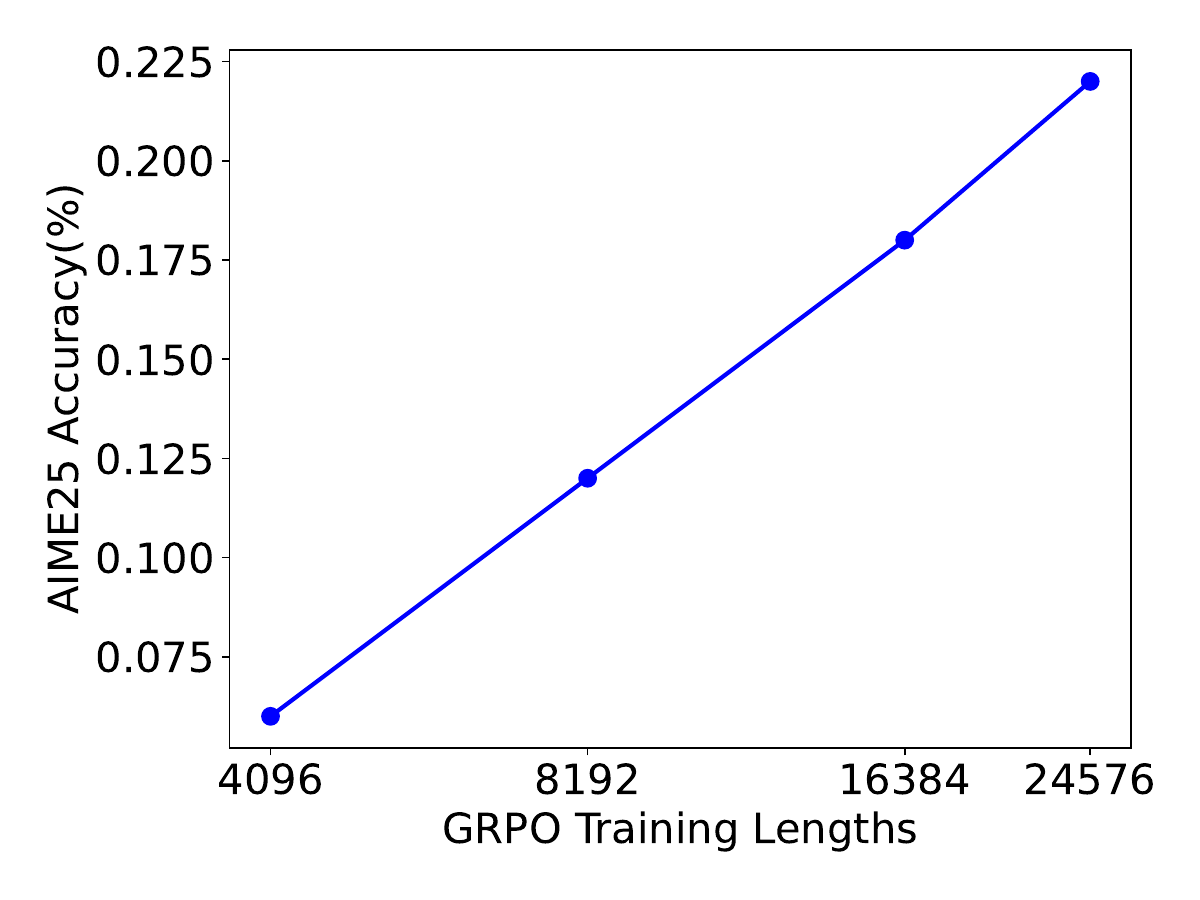}
  \caption{Pass@1 vs. maximum sequence length in GRPO training}
  \label{fig:rl-training-length}

  \vspace{0.1em} 

  \centering
  \begin{tabular}{lcc}
    \toprule
     & MATH500 & AIME24 \\
    \midrule
    Distill & 38 & 0 \\
    Distill + SFT(MATH) & 45 & 0 \\
    Distill + SFT(MATH) + SFT(Reason) & 74 & 22 \\
    Distill + SFT(MATH) + SFT(Reason) + RL & 82 & 28 \\
    \bottomrule
  \end{tabular}
  \captionof{table}{\ssmname{} Accuracy after each training stage on MATH500 and AIME24.}
  \label{tab:math_results_rotated}
\end{figure}

With more efficient models, we can increase the length of sequences used in RL training, resulting in improved performance. Empirically, we see this in Figure \ref{fig:rl-training-length}, which shows an increase in accuracy on AIME25 as we scale up the length of sequences generated when training with GRPO. Training with sequences of maximum length 4096 results in accuracy below 10\%, while allowing sequences up to length 24K boosts the accuracy up to 23\%. 


\textbf{MATH Accuracy at each training stage}

To identify which components of our training pipeline have the greatest impact on performance, we also evaluate intermediate versions of the model on MATH500~\citep{hendrycks2021measuringmathematicalproblemsolving} and AIME24~\citep{aime24}. The results of these evaluations are presented in Table \ref{tab:math_results_rotated}. Each step of the training pipeline provides a boost to performance, with particularly large gains from fine-tuning on solutions from reasoning models (+29\% on MATH500 and +17\% on AIME24).


\textbf{Direct Distillation from Reasoning Models}
We also attempted to distill from Deepseek-R1-Qwen-1.5B instead of Llama-3.2-3B. In this case, we did not SFT on OpenMathInstruct, and instead only SFT on the 10B reasoning data that we collected after distillation. We found that the distilled model's performance was poor (38\% and 3.3\% pass@1 accuracy on MATH500 and AIME24, resspectively). Our hypothesis for why this occurs is that 10B tokens is insufficient to effectively transfer reasoning skills from the transformer to Mamba. Although curating a high-quality reasoning dataset demands significant time and effort, we begin by leveraging the standard MATH distillation dataset from OpenMathInstruct~\citep{toshniwal2024openmathinstruct2acceleratingaimath} to first distill a strong MATH model. We then transform this MATH model into a reasoning model via SFT on the dedicated reasoning dataset. This approach achieves strong performance with a much smaller number of reasoning tokens.







\section{Conclusion}

In this paper, we introduced M1, a hybrid reasoning model built on the Mamba architecture, designed to address the scalability challenges of the Transformer models. We demonstrated effective techniques for distillation and finetuning to develop M1, which achieves mathematical reasoning performance comparable to state-of-the-art reasoning models of similar size. Notably, M1 delivers over 3x faster inference than similar-sized Transformer models, even when using the heavily optimized vLLM inference engine, particularly at large batch sizes. This improved efficiency can make the resource-intensive inference-time strategies, such as self-consistency, more practical. Our findings establish M1 as a strong alternative to Transformer-based architectures, paving the way for more efficient and high-performing reasoning models.




\bibliography{colm2025_conference}
\bibliographystyle{colm2025_conference}

\appendix
\section{Limitations and Future Work}

\paragraph{Speedup.}
Our current hybrid model is only 3× faster than a Transformer of the same size when serving inference with vLLM. Recently, NVIDIA introduced a new hybrid Mamba kernel\footnote{\url{https://github.com/NVIDIA/Megatron-LM/commit/b957578e76a921209ef873cbbd389114a4042542}}, which could further boost the speed of hybrid models. Additionally, our attention implementation in hybrid models does not yet leverage the optimizations available in vLLM. Integrating \ssmname{} into vLLM could further boost performance by taking advantage of these attention speedups.

\paragraph{Why do we not distill Qwen2.5 1.5B MATH model.} 
We considered using the Qwen2.5 1.5B MATH Instruct model as the distillation target in the first stage. However, we found that the cross entropy loss of the Qwen 1.5B MATH model on the OpenMATH Instruct dateset ~\citep{toshniwal2024openmathinstruct2acceleratingaimath} exceeded 1.8, which is much higher than that of the Llama models (0.5). This suggests that, to mimic the Qwen2.5 model, we need a dataset generated from a large Qwen2.5 series model rather than this one generated from the Llama models. Dataset curation from Qwen Math models goes beyond the scope of this work.

\paragraph{Improvement on RL training speed}
Recently, DeepSeek R1~\citep{deepseekai2025deepseekr1incentivizingreasoningcapability} showed that reinforcement learning (RL) is a key component in improving model reasoning performance during post-training. Since then, recent research has predominantly relied on reinforcement learning (RL) as a training paradigm for reasoning models. However, training with RL requires the efficient generation of long sequences. For example, in VeRL~\citep{sheng2024hybridflow}, the typical training batch size ranges from a few thousand to several thousand. DeepscaleR~\citep{deepscaler2025} also shows a significant accuracy boost when training RL with longer sequences, as it tends to enhance model performance by providing more steps for thorough reasoning. However, this shift towards reinforcement learning has resulted in the generation process becoming a significant bottleneck in reasoning model training, taking more than three times as long as the actor's weight update (forward + backward) according to the time profiling done for DeepscaleR~\citep{deepscaler2025}. This need for efficient generation in RL presents a significant challenge for transformer models, namely due to the heavy computational burden imposed by large key-value caches during generation, especially for large batch sizes. Given their generation speed advantages, linear RNN models may be better suited for scaling RL training.


\end{document}